\documentclass[11pt]{article}

\usepackage{jmlr2e}
\usepackage{lastpage}

\usepackage[margin=1in]{geometry}
\usepackage{csquotes}
\usepackage{epsfig}
\usepackage{color}
\usepackage{balance} 
\usepackage{algorithmic}
\usepackage{algorithm}
\usepackage{multirow}
\usepackage{graphicx}
\usepackage{amsmath}
\usepackage{amssymb}
\usepackage{amsfonts}
\usepackage{xspace}
\usepackage{url}
\usepackage{booktabs}
\usepackage{bbold}
\usepackage{changepage, dsfont}
\usepackage{pifont}
\usepackage{listings}
\usepackage[svgnames]{xcolor}
\usepackage{footmisc}

\newcommand{\IGNORE}[1]{}

\newcommand{\normaltilde}{\raise.17ex\hbox{$\scriptstyle\sim$}}

\jmlrheading{24}{2023}{1-\pageref{LastPage}}{2/22; Revised
4/23}{6/23}{22-0189}{Hussein Hazimeh, Rahul Mazumder, and Tim Nonet}

\ShortHeadings{L0Learn: A Scalable Package for Sparse Learning using L0 Regularization}{Hazimeh, Mazumder, and Nonet}
\firstpageno{1}

\begin{document}


\title{L0Learn: A Scalable Package for Sparse Learning using L0 Regularization}

\author{\name Hussein Hazimeh \email hazimeh@google.com \\
        \addr Google Research
        \AND
       \name Rahul Mazumder \email rahulmaz@mit.edu \\
       \addr Massachusetts Institute of Technology
       \AND 
        \name Tim Nonet \email tim.nonet@gmail.com \\
        \addr
               Massachusetts Institute of Technology}

\editor{Alexandre Gramfort}

\maketitle
\begin{abstract}
We present \texttt{L0Learn}: an open-source package for sparse linear regression and classification using $\ell_0$ regularization. \texttt{L0Learn} implements scalable, approximate algorithms, based on coordinate descent and local combinatorial optimization. The package is built using C++ and has user-friendly R and Python interfaces. 
\texttt{L0Learn} can address problems with millions of features, achieving competitive run times and statistical performance with state-of-the-art sparse learning packages. \texttt{L0Learn} is available on both CRAN and GitHub.\footnote{Links: \url{https://cran.r-project.org/package=L0Learn} and \url{https://github.com/hazimehh/L0Learn} \label{link:l0learn}}
\end{abstract}

\section{Introduction}
High-dimensional data is common in many important applications of machine learning, such as genomics and healthcare \citep{bycroft2018uk}. 
For statistical and interpretability reasons, 
it is desirable to learn linear models with sparse coefficients~\citep{hastie2015statistical}. 
Sparsity can be directly obtained using $\ell_0$ regularization, which controls the number of nonzero coefficients in the model. To illustrate this, let us consider the standard linear regression problem with a data matrix $X \in \mathbb{R}^{n \times p}$, regression coefficients $\beta \in \mathbb{R}^p$, and a response vector $y \in \mathbb{R}^{n}$. Using $\| \beta \|_0$ to denote the number of nonzero entries in $\beta$,
the $\ell_0$-regularized least-squares problem is given by:
\begin{align} \label{eq:l0_regression}
    \min_{\beta} ~~ \frac{1}{2} \| y - X \beta \|_2^2 + \lambda \| \beta \|_0,
\end{align}
where $\lambda 
\geq 0$ is a regularization parameter. Problem~\eqref{eq:l0_regression} is a fundamental problem in statistics and machine learning known as \textsl{best subset selection}, since it searches for a subset of the features that leads to the best fit (in terms of squared error) \citep{hocking1967selection}. 
Statistical optimality properties of estimator~\eqref{eq:l0_regression} have been discussed in~\cite{greenshtein2006best,raskutti2011minimax,zhang2014lower}. Problem \eqref{eq:l0_regression} is NP-hard and poses computational challenges. 
Various approaches have been proposed to approximate solutions to~\eqref{eq:l0_regression}. These include continuous proxies to the $\ell_0$ norm such as the $\ell_1$ norm \citep{tibshirani1996regression} and nonconvex penalties such as SCAD and MCP \citep{fan2001variable,mcp}. For these continuous penalties, specialized software packages have been developed, e.g., \texttt{glmnet} \citep{glmnet}, \texttt{ncvreg} \citep{ncvreg}, \texttt{sparsenet}~\citep{sparsenet}, and \texttt{picasso} \citep{picasso}.
Approaches for cardinality-constrained problems include greedy heuristics (e.g., stepwise methods), iterative hard thresholding (IHT) \citep{blumensath2008iterative}, BeSS~\citep{wen2020bess}, \texttt{abess} \citep{zhu2020polynomial}, among others.

Recently, there \textcolor{black}{have} been significant advances in computing globally optimal solutions to $\ell_0$-regularized problems, by using \textcolor{black}{mixed integer programming \citep{wolsey1999integer}}---see for example \citet{bestsubset,bertsimas2017sparse,hazimeh2022sparse}, and references therein. For example, $\ell_0$-regularized regression problems can be solved to optimality in minutes to hours for $p \sim 10^7$ when highly sparse solutions are desired \citep{hazimeh2022sparse}. 
In several applications however, high-quality approximate solutions for $\ell_0$-regularized problems may be more practical
than obtaining global optimality certificates.  
Our recent works~\citep{fastbestsubset,dedieu2021learning} propose fast, approximate algorithms for computing high-quality solutions to $\ell_0$-regularized problems, with running times comparable to that of fast $\ell_1$ solvers. Our algorithms are based on a combination of coordinate descent (CD) and local combinatorial optimization, for which we establish convergence guarantees. 
Local search often results in improved solution quality over using CD alone. Indeed, the experiments in \citet{fastbestsubset,dedieu2021learning} indicate that our algorithms can  outperform state-of-the-art methods based on $\ell_1$, MCP, SCAD, IHT, and others in terms of different statistical metrics (prediction, estimation, variable selection) and across a wide range of statistical settings. 

This paper introduces \texttt{L0Learn}: a package for $\ell_0$-regularized linear regression and classification. In addition to the $\ell_0$ penalty, \texttt{L0Learn} supports continuous
penalties such as the $\ell_1$ or squared $\ell_2$ norm. \texttt{L0Learn} implements highly optimized CD and local combinatorial optimization algorithms~\citep{fastbestsubset,dedieu2021learning} in C++, along with user-friendly R and Python interfaces to support fitting and visualizing models.

\section{Package Overview}

\textbf{Problem formulation.} In \texttt{L0Learn}, we consider a supervised learning setting with samples $\{ (x_i, y_i) \}_{i=1}^n$, where $x_i \in \mathbb{R}^{p}$ is the $i$-th feature vector and $y_i \in \mathbb{R}$ is the corresponding response. Let $L: \mathbb{R} \times \mathbb{R} \to \mathbb{R}$ be an associated loss function (such as squared error or logistic loss). \texttt{L0Learn} approximately minimizes the empirical risk, penalized with an $\ell_0 \ell_1$ or $\ell_0 \ell_2$ regularization.
Specifically, for a fixed $q \in \{1,2\}$, we compute approximate solutions to 
\begin{align} \label{eq:L0Learn}
    \min_{\beta_0, \beta} ~~ \sum_{i=1}^{n} L(y_i, \beta_0 +  x_i^T\beta) + \lambda \| \beta \|_0 + \gamma \| \beta \|_q^q,
\end{align}
where $\lambda$ and $\gamma$ are non-negative regularization parameters. While the $\ell_0$ penalty performs variable selection, the $\ell_q$ regularization induces shrinkage to help mitigate overfitting, especially in low-signal settings  \citep{mazumder2023subset,fastbestsubset}. \texttt{L0Learn}  supports regression using squared-error loss, classification using logistic and squared-hinge losses; and also accommodates box constraints on the $\beta_i$s. 

\textbf{Overview of the algorithms.} \texttt{L0Learn} 
uses a combination of (i) cyclic CD and (ii) local combinatorial optimization. The choice of CD is inspired by its strong performance in sparse learning with continuous penalties~\citep{glmnet,sparsenet}. Standard convergence results for cyclic CD, e.g., \citet{Tseng01}, do not apply for the discontinuous objective in~\eqref{eq:L0Learn}. 
Hence, in~\citet{fastbestsubset}, we show that a variant of cyclic CD converges to a local minimizer of Problem~\eqref{eq:L0Learn}.  
Given a CD solution $(\beta_0, \beta)$, the local combinatorial optimization algorithm searches a local neighborhood for solutions with a better objective.\footnote{We implement the case where the neighborhood consists of all solutions obtained by removing one variable from the support of $\beta$ and adding another previously zero variable to the support.} After finding a new improved solution using local search, we run CD with the new solution in an attempt to further improve the current solution. We keep iterating between local search and CD, until convergence---see \citet{fastbestsubset} for details.

 \textbf{Efficient computation of the regularization path.} \texttt{L0Learn} solves \eqref{eq:L0Learn} over a grid of regularization parameters. We use computational schemes such as warm starts, active sets, greedy cyclic order for CD, and correlation screening \citep{fastbestsubset}, to speed up the algorithm.
Due to the nature of the $\ell_0$ penalty, different values of $\lambda$ in \eqref{eq:L0Learn} can lead to the same solution. Thus, to avoid duplicate solutions and unnecessary computations, we develop a new method that computes a data-dependent grid of $\lambda$s, which is guaranteed to avoid duplicate solutions---see \citet{fastbestsubset} for details.

 \textbf{Implementation and development.} All the algorithms in the  package are implemented in C++, along with high-level R and Python interfaces. For linear algebra operations, we rely on the \texttt{Armadillo} library \citep{sanderson2016armadillo}, which is accelerated by Basic Linear Algebra Subprograms (BLAS) \citep{lawson1979basic}. The R interface is integrated with C++ using \texttt{RcppArmadillo} \citep{eddelbuettel2014rcpparmadillo}. The Python interface is integrated with C++ using \texttt{carma} \citep{Urlus_CARMA_bidirectional_conversions_2023} and \texttt{pybind11} \citep{pybind11}. \texttt{L0Learn} supports both dense and sparse data matrices. Sparse matrices generally speed up computation and reduce memory requirements. In all functions that require the data matrix as an input argument, we use function templates that accept a generic matrix type. Thus, exactly the same code is used for both dense and sparse matrices, but each matrix type uses specialized linear algebra implementations. 

For development, we use continuous integration based on Travis CI to build and test the package. Our unit and integration tests are primarily implemented in R based on \texttt{testthat} \citep{wickham2011testthat} and have a coverage of {97\%} (as measured by \texttt{covr}). \texttt{L0Learn} achieved the highest rating (A) for code quality by the code analysis tool Codacy (\url{www.codacy.com}). We have additional CI checks that use Github Actions for our Python bindings to ensure the Python bindings interact with the C++ library identically to the R bindings. 

\begin{figure}[tp]
    \begin{lstlisting}[language=R]
    # Assume the data matrix (x) and response (y) have been loaded
    fit <- L0Learn.fit(x, y, penalty="L0") # Fit an L0 regularized regression model
    plot(fit) # Plot the regularization path
    cv_fit <- L0Learn.cvfit(x, y, penalty="L0", nFolds=5) # 5-fold cross validation
    plot(cv_fit) # Plot the cross-validation error
    \end{lstlisting}
    \caption{Simple examples of using \texttt{L0Learn} in R.}
    \label{fig:example_code}
    \vspace{-0.6cm}
\end{figure}
\section{Usage and Documentation}
\texttt{L0Learn} can be installed in R by executing  \texttt{install.packages("L0Learn")} and can be installed in Python by executing \texttt{pip install l0learn}. It is supported on Linux, macOS, and Windows. In Figure \ref{fig:example_code}, we provide simple examples of how to use \texttt{L0Learn} in R. 
More elaborate examples and a full API documentation can be found in \texttt{L0Learn}'s Vignette and Reference Manual, which are available on \texttt{L0Learn}'s main pages\footref{link:l0learn}. Similar examples and API documentation are available in Python as well.

\section{Experiments} 
\textcolor{black}{Our main goal in these experiments is to compare the running time of \texttt{L0Learn} with similar toolkits, designed for sparse learning problems. Specifically, we compare with \texttt{glmnet}, \texttt{ncvreg}, \texttt{picasso}, and \texttt{abess}. For space constraints, we focus on linear regression, and refer the reader to~\citet{dedieu2021learning} for sparse classification experiments. Our experiments also shed some light on the statistical performance of the different approaches: for in-depth studies of statistical properties, see  \citet{fastbestsubset,hastie2020best,mazumder2023subset}. We note that there are modern approaches for solving $\ell_0$-regularized problems to global optimality e.g., \citet{bertsimas2017sparse,hazimeh2022sparse,hazimeh2023grouped}. However, due to their focus on optimality certification, they are usually (much) slower than the competing toolkits we present here \citep{hazimeh2022sparse}.}

\textbf{Setup.} 
\textcolor{black}{Following~\cite{fastbestsubset}, we consider synthetic data as per a linear regression model under the fixed design setting (exponential correlation model with $\rho=0.3$). We take $n=1000$, $k=50$, signal-to-noise ratio (SNR)\footnote{Similar to \cite{bestsubset}, SNR := Var($X \beta^{*}$)/$\sigma^2$.} to be 5}
and vary $p \in \{ 10^3, 10^4, 10^5 \}$.
In \texttt{L0Learn}, we used the default CD algorithm with the $\ell_0 \ell_2$ penalty. In \texttt{picasso}, we used  $\ell_1$ regularization and changed the convergence threshold (prec) to $10^{-10}$ so that its solutions roughly match those of \texttt{glmnet}. In \texttt{ncvreg}, we used the (default) MCP penalty. All competing methods are tuned to minimize MSE on a validation set with the same size as the training set. In \texttt{L0Learn}, \texttt{ncvreg}, and \texttt{abess}, we tune over a two-dimensional grid consisting of $100$ $\lambda$ values (chosen automatically by the toolkits) and $100$ $\gamma$ values (in the range $[10^{-2}, 10^{2}]$ for \texttt{L0Learn}, $[1.5, 10^3]$ for \texttt{ncvreg}, $[1, 10^3]$ for \texttt{abess}\footnote{We found out that \texttt{abess} requires a different range for $\gamma$  (the parameter of the squared $\ell_2$ reguralizer)  to perform well. This is expected since \texttt{abess} considers a different, cardinality-constrained formulation.}). Experiments were performed on a Linux c5n.2xlarge EC2 instance running R 4.0.2.

\textbf{Metrics.} We report the running time in seconds for computing a path of 100 solutions. \textcolor{black}{(L0Learn, \texttt{abess}, and \texttt{ncvreg} require additional time for tuning their second parameter~$\gamma$)}. In addition, given an estimator $\hat{\beta}$, we compute the following statistical metrics: (i)~prediction error (PE) given by $\| X \hat{\beta} - X \beta^{*}  \|_2 / \| X \beta^{*} \|_2$, (ii)~the number of false positives (FP), i.e., the nonzero variables in $\hat{\beta}$ that are not in $\beta^{*}$, and (iii) the support size (SS).

\textbf{Results.} In Table \ref{table:main}, we report the metrics averaged over $10$ repetitions. The results indicate that \texttt{L0Learn} achieves best-in-class run times and statistical performance. \textcolor{black}{While we only show results on synthetic data sets due to space constraints, \texttt{L0Learn} is useful and can lead to good improvements on real data sets, as demonstrated in many studies, including  \citet{fastbestsubset,o2021sparse,cao2021reconstruction,li2023integration}.}

\begin{table}[t] 
\centering
\textcolor{black}{
\scalebox{1.0}{
\resizebox{0.99\columnwidth}{!}{%
\begin{tabular}{l|cccc|cccc|cccc|}
\cline{2-13}
                              & \multicolumn{4}{c|}{$p=10^3$}                                                                                     & \multicolumn{4}{c|}{$p=10^4$}                                                                                     & \multicolumn{4}{c|}{$p=10^5$}                                                                                     \\ \cline{2-13} 
                              & \multicolumn{1}{c}{Time} & \multicolumn{1}{c}{PE$\times 10^2$} & \multicolumn{1}{c}{FP} & \multicolumn{1}{c|}{SS} & \multicolumn{1}{c}{Time} & \multicolumn{1}{c}{PE$\times 10^2$} & \multicolumn{1}{c}{FP} & \multicolumn{1}{c|}{SS} & \multicolumn{1}{c}{Time} & \multicolumn{1}{c}{PE$\times 10^2$} & \multicolumn{1}{c}{FP} & \multicolumn{1}{c|}{SS} \\ \hline
\multicolumn{1}{|l|}{L0Learn} & \textbf{0.09 (0.01)}              & \textbf{9.4 (1.2)}                          & \textbf{0 (0)}                  & \textbf{50 (0)}                  & \textbf{0.49 (0.0)}               & \textbf{9.4 (0.8)}                           & \textbf{0 (0)}                  & \textbf{50 (0)}                  & \textbf{4.4 (0.4)}                & \textbf{9.5 (1.1)}                           & \textbf{0 (0)}                  & \textbf{50 (0)}                  \\
\multicolumn{1}{|l|}{glmnet}  & 0.55 (0.02)              & 19.8 (1.0)                          & 154 (22)               & 204 (22)                & 0.94 (0.0)               & 26.4 (0.9)                          & 300 (22)               & 350 (22)                & 8.0 (0.1)                & 32.3 (1.5)                          & 485 (38)               & 535 (38)                \\
\multicolumn{1}{|l|}{picasso} & 1.46 (.15)               & 19.8 (1.0)                          & 157 (24)               & 207 (24)                & 2.92 (0.0)               & 26.4 (0.9)                          & 303 (22)               & 353 (22)                & 15.5 (0.2)               & 32.3 (1.5)                          & 483 (35)               & 533 (35)                \\
\multicolumn{1}{|l|}{abess}   & 0.43 (0.04)              & 11.0 (1.8)                          & 3.9 (3)                & 53.9 (3)                & 7.21 (0.1)               & 16.2 (7.8)                          & 2 (0)                  & 52 (0)                  & -                        & -                                   & -                      & -                       \\
\multicolumn{1}{|l|}{ncvreg}  & 1.35 (0.22)              & \textbf{9.4 (1.2)}                           & 2 (5)                  & 52 (5)                  & 3.74 (0.4)               & \textbf{9.4 (0.7)}                           & 4 (7)                  & 54 (7)                  & 19.7 (0.6)               & 9.6 (1.0)                           & 1 (1)                  & 51 (1)                  \\ \hline
\end{tabular}%
}}
\caption{\small{The mean and standard error of the running time (s), prediction error (PE), number of false positives (FP), and support size (SS). A dash  indicates failure due to memory issues.}}
\label{table:main}
}
\end{table}

\section{Conclusion}
We introduced \texttt{L0Learn}: a scalable package for $\ell_0$-regularized regression and classification. The package is implemented in C++ along with R and Python interfaces. It offers two approximate algorithms: a fast coordinate descent-based method and a local combinatorial search algorithm that helps in improving solution quality. Our experiments indicate that \texttt{L0Learn} is highly scalable and can outperform popular sparse learning toolkits in important high-dimensional settings. The package has been successfully used in a variety of applications in healthcare and genetics; and has also proven to be effective in warm starting exact $\ell_0$ regularization solvers  \citep{hazimeh2022sparse}. As a future direction, it would be interesting to study the performance of CD and local combinatorial search in ($\ell_0$-based) pruning of neural networks \citep{benbaki2023fast}.

\section*{Acknowledgements}
We acknowledge research support from the Office of Naval Research and National Science Foundation.

\clearpage
\bibliography{ref}

\end{document}